\documentclass[final]{cvpr}

\usepackage{times}
\usepackage{epsfig}
\usepackage{graphicx}
\usepackage{amsmath}
\usepackage{amssymb}
\usepackage{xparse}
\usepackage{multirow}
\usepackage[dvipsnames]{xcolor}

\usepackage[ruled,vlined]{algorithm2e}

\usepackage[pagebackref=true,breaklinks=true,letterpaper=true,colorlinks,bookmarks=false]{hyperref}

\begin{document}

\newcommand{\appref}[1]{Appendix~\ref{#1}}
\newcommand{\ogvqa}{VQA v2\xspace}
\newcommand{\rulesep}{\unskip\ \vrule\ }

\title{Separating Skills and Concepts for Novel Visual Question Answering}

\author{
Spencer Whitehead$^{1}$\thanks{Work was partly done as an intern at the MIT-IBM Watson AI Lab.} , Hui Wu$^{2}$, Heng Ji$^{1}$, Rogerio Feris$^{2}$, Kate Saenko$^{2,3}$\\
$^{1}$UIUC \quad $^{2}$MIT-IBM Watson AI Lab, IBM Research \quad $^{3}$Boston University\\
{\tt\small \{srw5,hengji\}@illinois.edu},
{\tt\small \{wuhu,rsferis\}@us.ibm.com},
{\tt\small saenko@bu.edu}
}

\maketitle
\thispagestyle{empty}
\pagestyle{empty}

\begin{abstract}
Generalization to out-of-distribution data has been a problem for Visual Question Answering (VQA) models.
To measure generalization to novel questions, we propose to separate them into ``skills'' and ``concepts''.
``Skills'' are visual tasks, such as counting or attribute recognition, and are applied to ``concepts'' mentioned in the question, such as objects and people.
VQA methods should be able to compose skills and concepts in novel ways, regardless of whether the specific composition has been seen in training, yet we demonstrate that existing models have much to improve upon towards handling new compositions.
We present a novel method for learning to compose skills and concepts that separates these two factors implicitly within a model by learning grounded concept representations and disentangling the encoding of skills from that of concepts. We enforce these properties with a novel contrastive learning procedure that does not rely on external annotations and can be learned from unlabeled image-question pairs.
Experiments demonstrate the effectiveness of our approach for improving compositional and grounding performance.\footnote{Code: {\scriptsize \url{https://github.com/SpencerWhitehead/novelvqa}}}
\end{abstract}

\section{Introduction}

\begin{figure}[t!]
\begin{center}
\includegraphics[width=\linewidth]{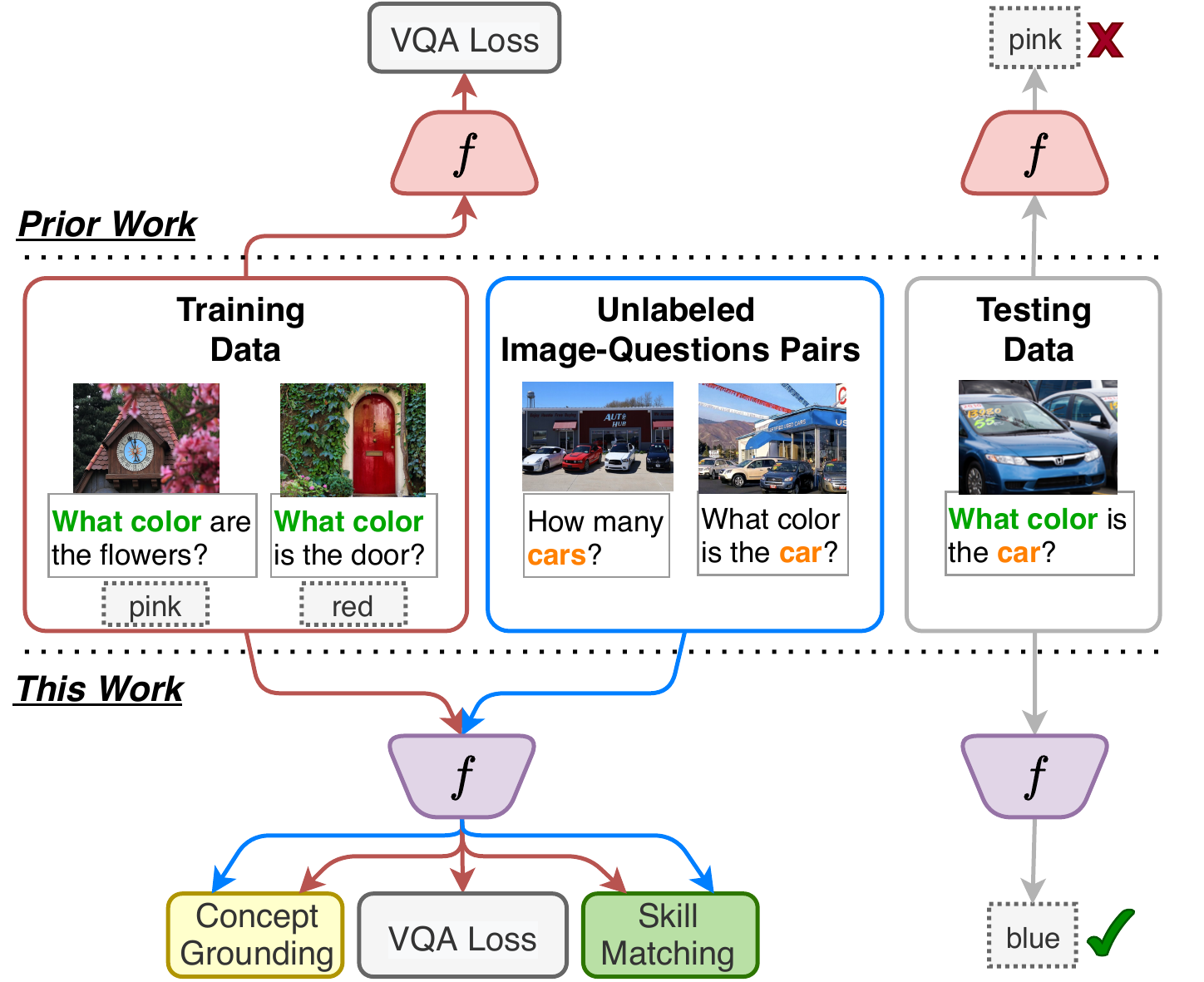}
\end{center}
\caption{We propose a new view of compositionality in VQA that explores the ability to answer questions about unseen compositions of skills ({\sl e.g.,} \textcolor{Green}{color}) and concepts ({\sl e.g.,} \textcolor{orange}{car}). We present a method that learns to separate skills and concepts that can utilize both labeled and unlabeled image-question pairs in order to generalize to novel questions with new skill-concept compositions and new concepts.}
\label{fig:teaser}
\end{figure}

When humans answer questions, such as in Visual Question Answering (VQA), we first interpret the question, dissecting its content into parts (like concepts, 
relations, actions, question types), and then we select and execute the \emph{skill} (or plan/program) necessary to produce an answer based on this information and the relevant knowledge base ({\sl e.g.,} the image)~\cite{hudson2018compositional,lehnert1977human,weston2015towards,skillsvqa2020}.
The skills needed to produce an answer are general and can be applied to (composed with) many types of question-specific content.
For example, if one can answer questions about ``\textit{colors}'' for a variety of objects as well as recognize and answer questions about ``\textit{cars}'', then questions like ``\textit{What color is the car?}'' should be straightforward to answer even if this specific composition has yet to be seen (\figref{fig:teaser}).
This ability of seamlessly adapting and composing conceptual representations with skills is crucial to demonstrating true understanding of VQA and learning to generalize from less labeled data.

Compositionality is recognized as one of the essential properties of human cognition~\cite{lake2017building}, but more research is still needed on incorporating compositionality into models and developing data-efficient, generalizable systems.
While much progress has been made to achieve better performance on standard VQA test benchmarks~\cite{anderson2018butd,goyal2017vqav2,kim2018bilinear,yu2019mcan}, most state-of-the-art models are still designed without any notion of built-in compositionality and tend to entangle skills and concepts in their learned representations.
Some previous work has studied the lack of generalization ability of VQA models, and evaluated models using test splits with different answer distributions from the training data.
However, this measurement only indirectly addresses the central issue (lack of compositionality), which manifests itself as poor generalization and over-reliance on language priors~\cite{agrawal2018vqacp,ramakrishnan2018}.

To address these issues, our first contribution is a new view of
VQA compositionality, called \emph{skill-concept composition}, and a new evaluation setting that directly targets how VQA models can generalize to novel compositions of skills and concepts.
This view is motivated by our observation that, to answer a natural question on real images requires the understanding of two distinct elements: 1) the visual concept referred to by the question; and 2) what information we need to extract from the referred concept.
We elucidate this in \secref{sec:paradigm} and evaluate a number of VQA architectures using this setting and demonstrate that the existing models have much to improve upon to answer novel questions.

We propose a novel approach to improve generalization that utilizes contrastive learning to separate skills and concepts within the internal representations of a model, while jointly learning to answer questions.
We use grounding as a proxy to separate concepts so that the model learns to identify a concept in both the question and image, regardless of the specific context.
Akin to weakly supervised grounding~\cite{akbari2019multi,gupta2020contrastive}, we train the model to recover a concept mentioned in a given image-question pair by contrasting the multi-modal representation of the masked concept word to the multi-modal representations of words in other questions.
We utilize a new way to curate positive and negative examples for the contrastive loss so that the model learns to predict the concept based on relevant visual information rather than using superficial contextual cues.
Additionally, our approach learns to separate skills from concepts by contrasting question representations that have the same or different skills.
These properties are learned jointly alongside the VQA objective, on top of state-of-the-art models, and are generalizable to new architectures.

Some advantages of our approach are:
1) We learn grounding in a self-supervised manner using the VQA data \emph{alone}, without external annotations.
This is in contrast to previous approaches with similar goals that incur large expenses due to annotation requirements~\cite{selvaraju2019vqahint,wu2019self}.
2) Our method does not rely on answer labels to learn skill-concept separation, so we are able to use \emph{unlabeled} image-question pairs to learn these properties.
Consequently, we are able to acquire new concepts and learn to answer questions about them without having labeled data with these concepts, which is pivotal for generalizing to a new domain or novel instances.
Moreover, we focus on data-efficient methods and do not use prodigious amounts of data external to VQA, like pre-training approaches~\cite{chen2020uniter,lu2019vilbert,tan2019lxmert}, which is expensive to obtain and can require prior knowledge of the domain and/or concepts in order to perform well~\cite{hendricks2021decoupling,singh2020pretrainright}.

Our main contributions in this paper are:
1) We present a novel view and evaluation setting for compositionality in VQA, called \emph{skill-concept composition}, which enables a more direct and interpretable evaluation of VQA models on real-image question answering.
2) We propose a novel contrastive learning approach, which combines the supervised VQA objective with self-supervised learning, to achieve skill-concept disentanglement at no additional annotation cost.
3) Our approach shows significant improvements over existing models on novel skill-concept compositions as well as generalization to unlabeled image-question pairs containing unseen concepts.

\begin{figure*}[!ht]
\begin{center}
\includegraphics[height=.37\textwidth]{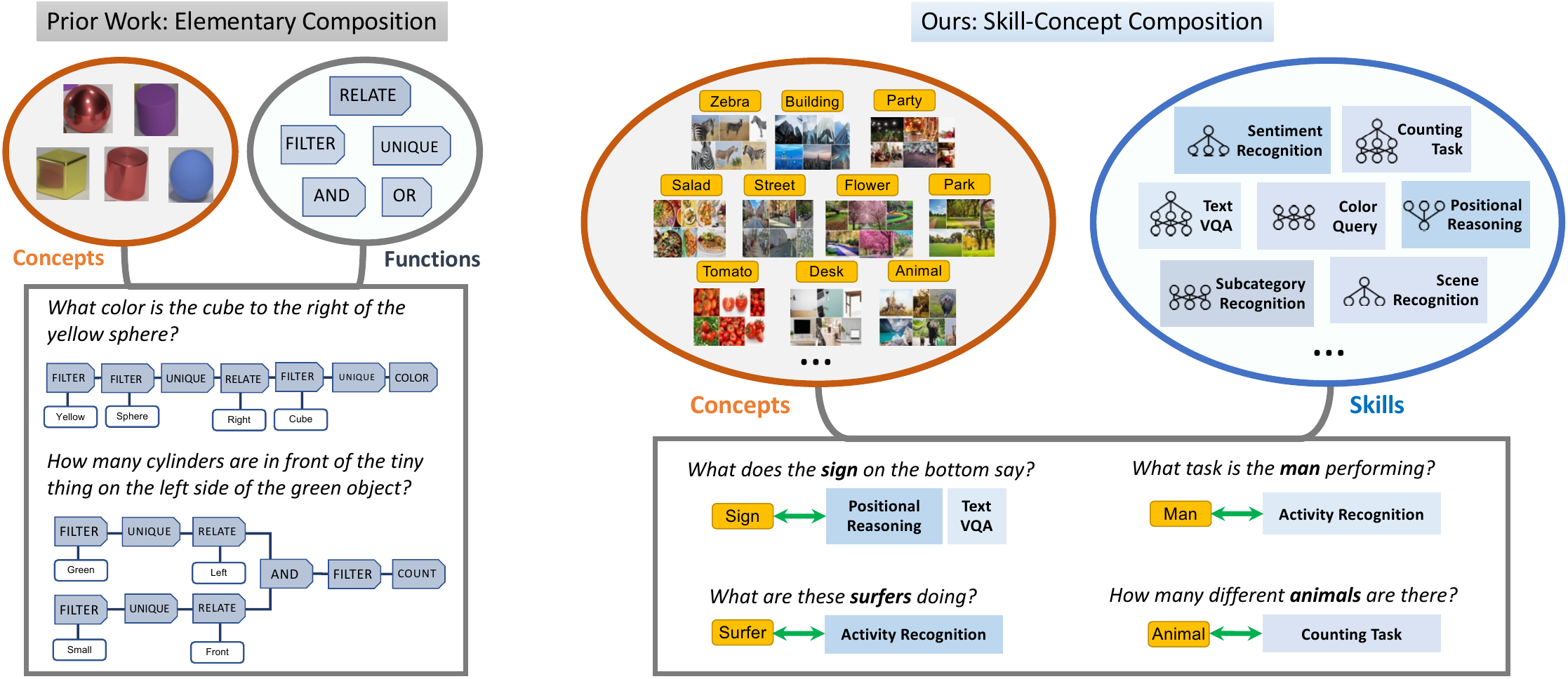}
\end{center}
\caption{Illustration of skill-concept composition, a new view and evaluation setting for compositionality in VQA.
}
\label{fig:separation}
\end{figure*}

\section{Related Work}

\noindent\textbf{VQA and Evaluations.}
Much progress has been made towards VQA~\cite{anderson2018butd,kim2018bilinear,li2020oscar,tan2019lxmert}.
Large-scale pre-trained transformers~\cite{vaswani2017attention}, inspired by BERT~\cite{devlin2019bert}, have recently become prevalent~\cite{chen2020uniter,huang2020pixel,li2019unicoder,li2019visualbert,li2020oscar,lu2019vilbert,su2019vl,tan2019lxmert,yu2019mcan}.
Progress is often measured by the widely adopted human-annotated \ogvqa~\cite{goyal2017vqav2} benchmark as well as other synthetic benchmarks~\cite{hudson2019gqa,johnson2017clevr,kafle2017analysis}.
Various datasets for reducing biases have been presented~\cite{agrawal2018vqacp,goyal2017vqav2}, including VQA-CP~\cite{agrawal2018vqacp} that creates distinct distributions of question prefixes and answers between the train and validation splits of \ogvqa.
Another dataset~\cite{kafle2017analysis} evaluates on a conglomeration of data from VQA v1~\cite{antol2015vqa} and synthetic data generated from question templates and image annotations, which they break down by vision task.
The basic premise of considering questions by task is similar to our ``skills'', but our setting examines compositional generalization of skills and concepts, where we evaluate on unseen compositions of skills and concepts.
Existing work for evaluating generalization~\cite{agrawal2018vqacp,kafle2017analysis} does not explore generalization to such novel compositions.

\noindent\textbf{Compositionality and VQA.} 
CLEVR~\cite{johnson2017clevr} and GQA~\cite{hudson2019gqa} 
are two VQA datasets that have received interest in recent years. Both offer \emph{compositional questions}, which means that the questions involve various relational chains ({\sl e.g.,} ``\textit{What color is the apple to the left of the bowl on the table?}'').
While GQA does not focus on novel compositions, CLEVR does investigate novel compositions of attributes and objects, where, for example, models see cubes of certain colors and cylinders of other colors during training, then the cubes and cylinders have their colors swapped in testing.
This is analogous to our setting, although we propose to investigate skill-concept compositions and we experiment with natural questions about real images.
Other efforts create compositional models to handle the relational reasoning chains~\cite{hu2018explainable,hu2017n2nnmn,hudson2018compositional,hudson2019learning,shi2019xnm}.
Our approach implicitly learns compositional capabilities within state-of-the-art multi-modal transformer architectures, unlike these methods that do so explicitly. We also provide results on how well current compositional models ({\sl i.e.,} neural module networks) can generalize to novel questions.

\noindent\textbf{Grounding Visual Concepts.}
Visual grounding is typically studied on image-caption pairs.
Previous work on visual grounding often learns grounding in a weakly supervised manner~\cite{akbari2019multi,datta2019align2ground,fang2015captions,gupta2020contrastive}, and sometimes leverages contrastive learning techniques~\cite{akbari2019multi,gupta2020contrastive}.
However, grounding concepts that appear in VQA questions has been less studied in previous work, which we explore in our paper.
One direction that is somewhat related to visual grounding in VQA is learning VQA models which have interpretable visual attentions~\cite{selvaraju2019vqahint,wu2019self}.
However, these approaches usually rely on human annotations of the most influential regions for VQA answer prediction~\cite{vqahat,huk2018multimodal}, which are expensive to obtain and do not directly associate concept mentions with visual regions.
Our work learns to ground concepts mentioned in questions, which facilitates compositional VQA modeling, and does so without using any additional annotations.

\begin{figure*}[t]
\small
\begin{center}
\includegraphics[height=0.35\linewidth]{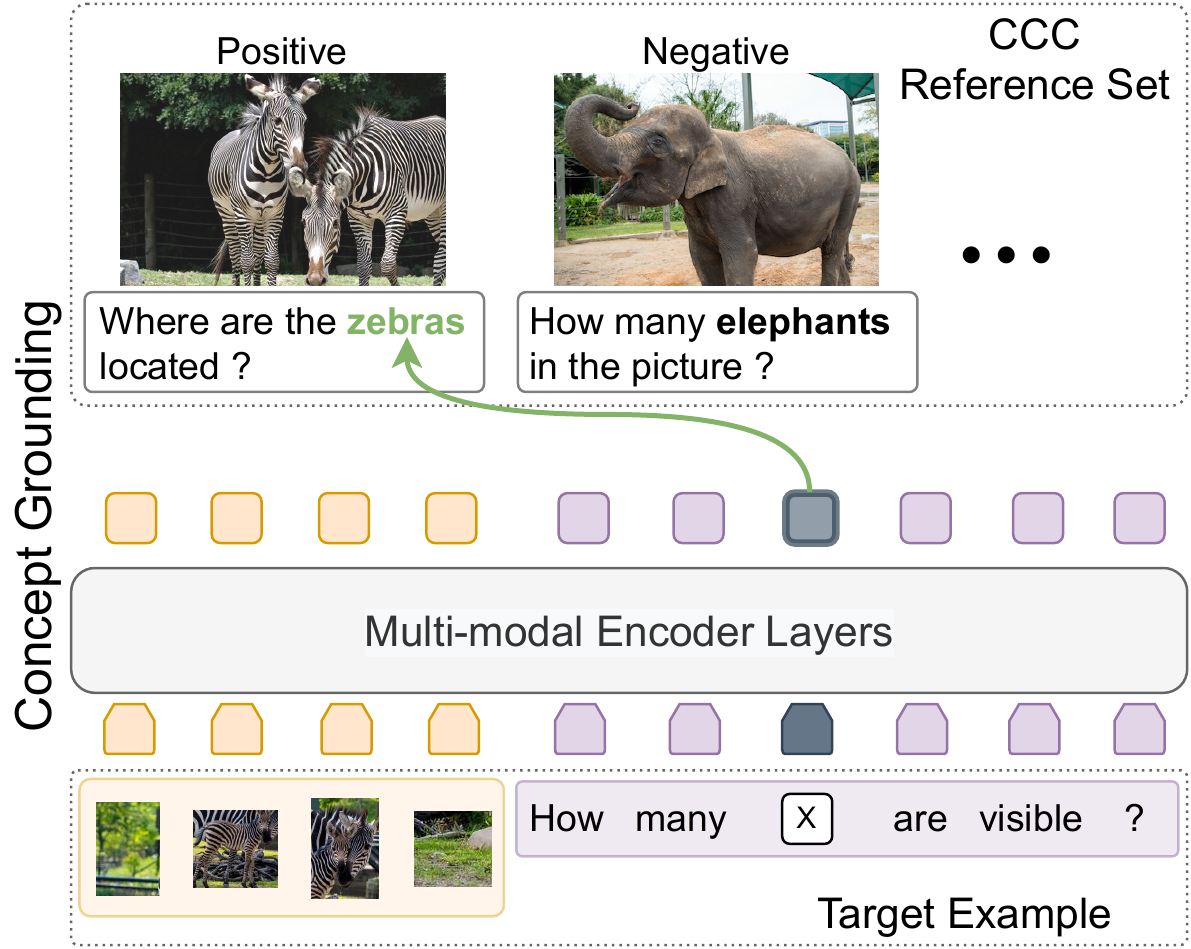}
\quad
\rulesep
\includegraphics[height=0.35\linewidth]{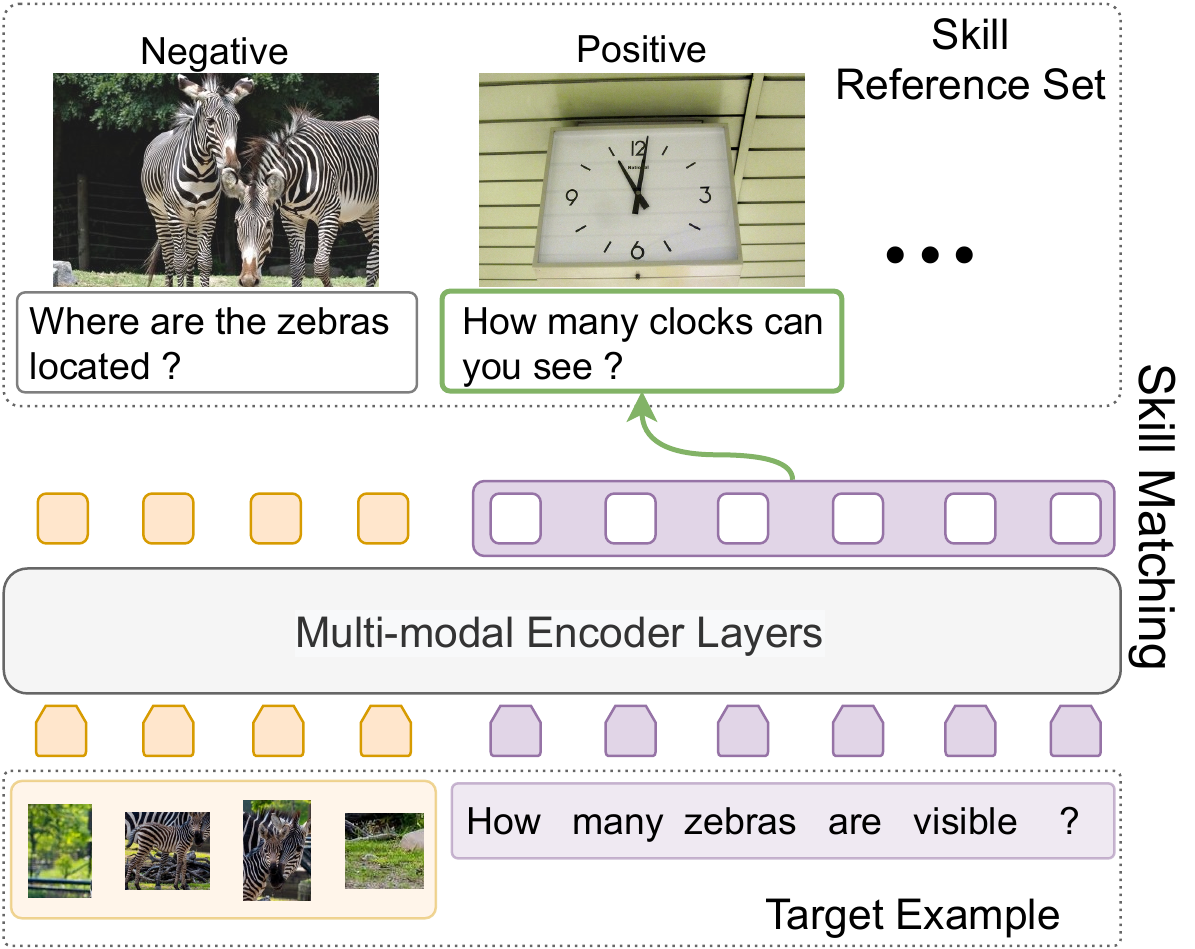}
\end{center}
\caption{Overview of our approach. Left: We learn to ground concept representations by contrasting the multi-modal representations of a masked concept token in the target example and words in other questions. Right: We encode skills in the summary representations of the question by contrasting with summary representations of other questions with the same (positive) or different (negative) skills.}
\label{fig:architecture}
\end{figure*}

\section{Skill-Concept Composition in VQA}\label{sec:paradigm}

We propose a novel, compositional view of VQA, called \emph{skill-concept composition} (\figref{fig:separation}).
\emph{Concepts} are objects and other visually-grounded words or phrases.
By \emph{skills}, we refer to the collection of high-level vision understanding processes involved in answering common questions about real-world images.
These skills operate on concepts and vary in terms of input/output representation complexity and the necessary reasoning processes.
Our taxonomy of these skills is extracted from annotating a subset of the VQA v2 questions as well as taking inspiration from prior work on VQA.\footnote{Please refer to \appref{appendix:skillconcept} for a complete skill list.}
Skills are generally standalone from each other and have been studied independently in the VQA literature ({\sl e.g.,} TextVQA~\cite{singh2019}, positional reasoning~\cite{hudson2019gqa}, or counting~\cite{acharya2019}).

We make an important yet intuitive observation about these VQA skills: to answer a question, it often requires the application of only a small number of skills (most often one) to one or more concepts in the image (\figref{fig:separation}).
This observation provides an interpretable view of a model's generalization ability to out-of-distribution data: a model should learn that a skill is a separable process that can be applied to different concepts, and that the prediction process should not be tied to specific concepts co-occurring with this skill during training.
This explicit notion of skill-concept separation underlies the contributions of this paper, including a new novel-VQA evaluation method which we will introduce next, as well as a new framework to weakly learn VQA models that can answer novel questions (\secref{sec:method}).

\noindent\textbf{Novel-VQA Evaluation.}  
While conceptually intuitive, this skill-concept view offers natural ways to guide the evaluation of VQA models in terms of out-of-distribution data.
In our experiments, we evaluate two novel-VQA settings: 1) answering questions on novel compositions of skills and concepts; 2) answering questions about concepts for which the model has not seen any answers before.

\noindent\textbf{Comparison to Existing Evaluations.}
Our evaluation protocol is different from existing VQA benchmarks that also aim to measure VQA models' generalization ability.
VQA-CP~\cite{agrawal2018vqacp} builds train-test splits from \ogvqa/v1~\cite{antol2015vqa,goyal2017vqav2} with distinct answer distributions by greedily dividing the questions based on their annotated question types ({\sl i.e.,} first few words in the question: ``\textit{how many}'', ``\textit{is the}'',...) and answers, but this does not capture skill-concept compositions because these question types do not necessarily correspond to skills ({\sl e.g.,} ``\textit{Is the dog waiting?}'' requires action recognition and ``\textit{Is the sky blue?}'' requires color recognition yet both have question type ``\textit{is the}''), and the same skill-concept composition can appear in training and testing, which violates our novel-VQA setting.
TDIUC~\cite{kafle2017analysis} evaluates VQA accuracy on different categories of task types, without regard for the concepts in the questions.
\cite{teney2016zeroshotvqa} creates testing splits such that at least one word of a question is unseen during training, which does not consider skills.
None of these benchmarks directly address and evaluate skill-concept compositionality like our evaluation protocol.

\noindent\textbf{Skill-Concept vs.\ Elementary Compositions.}
Existing compositional evaluations primarily define compositions as relational reasoning chains associated with questions~\cite{hudson2019gqa,johnson2017clevr} (\figref{fig:separation}), which are suited for learning programs of elementary operations to answer questions.
There are two main issues with applying this existing compositional view to real-image question answering.
First, the concepts and their attributes are over-simplified and not representative of the diverse visual presentations in the real world.
Second, the kinds of compositional questions in these synthetic datasets rarely appear in natural questions about real-world images.
Our proposed \emph{skill-concept composition} view is more applicable to real-world VQA and, as a result, can better represent the capabilities that people care about in real-image question answering.

\section{Approach}\label{sec:method}

\noindent\textbf{Preliminaries.}
We assume that we are given a partially labeled dataset of tuples with image $I$, question $Q$, and answer labels $A$, where $(I^{a}, Q^{a}, A^{a})\in\mathcal{D}^{a}$ has labels and $(I^{u}, Q^{u})\in\mathcal{D}^{u}$ does not.
Typically, models are trained with a VQA loss~\cite{teney2018tips} using the labeled dataset, $\mathcal{D}^{a}$.
Given an example $(I, Q)$, image region features, $g_v(I) = \{v_1,...,v_M\}$, and question token embeddings, $g_w(Q) = \{x_1,...,x_N\}$, are extracted and input to a multi-modal encoder to produce multi-modal representations of both modalities, $f(g_v(I), g_w(Q)) = (\{z_m\}_{m=1}^{M}, \{h_i\}_{i=1}^{N})$, where $z_m$ and $h_i$ are the image and text multi-modal representations, respectively.
An answer is predicted by pooling the encoded representations to a single representation (or using a \textsc{CLS} token as input~\cite{devlin2019bert}), which is then input to a softmax output layer.
We build upon this basic VQA setup to learn skill-concept separation, and uniquely take advantage of both labeled and unlabeled data.

\noindent\textbf{Overview.}
We aim to learn separable skills and concepts, such that we can compose them to answer novel questions.
To do so, the model should recognize that concepts mentioned in the question are manifested by their appearances in the image ({\sl i.e.,} grounding) and that skills should be identifiable regardless of the concepts in the question or image.
Gathering supervision for identifying concepts in the question, grounding them in the image, and labeling questions with skills would be very costly.
Therefore, we propose to learn skill-concept separation in a self-supervised manner using contrastive learning~\cite{chen2020nce,oord2018representation}.
Illustrated in \figref{fig:architecture}, we train the model with two additional contrastive objectives jointly with the VQA objective: \emph{concept grounding} (\secref{sec:conceptground}), which learns grounded concept representations, and \emph{skill matching} (\secref{sec:skillmatch}), which encodes concept-agnostic representations of skills.
For each of our objectives, the model is presented with a \emph{target example} and a \emph{reference set} of positive and negative examples sampled from carefully curated \emph{candidate references}.
Each objective trains the model to make the representation from the target example similar to those of the positive ones.
We expound our training procedure for learning these objectives jointly with VQA in \secref{sec:train_procedure}.
In the following, for brevity, we put specific details of functions/settings in \appref{appendix:approach_details}.

\subsection{Concept Grounding}\label{sec:conceptground}

To learn grounded representations of concepts, we mask the concept mention from the target question and then train the model to recover this concept mention, using the multi-modal contextual information, by pointing to the same concept mention in examples in the reference set (\figref{fig:architecture}).

\noindent\textbf{Concept Discovery.}
We first identify the concept words that can be grounded in images.
While this can be done with different methods~\cite{dodge2012detecting,hessel2018vlwordconcrete,kehat2017vlwordconcrete}, we simply use heuristics.
We use POS tagging and lemmatization~\cite{spacy2} to identify the 400 most frequent nouns in \ogvqa and then we filter out concepts that cannot be grounded ({\sl e.g.,} ``\textit{time}'').
For a given question, $Q$, we want to find examples that have co-occurring mentions of a concept and the appearance of that concept in the image.
It is likely that if a question about an image mentions a concept, then that concept may appear in the image~\cite{ganju2017what}.
Therefore, we identify the set of questions that mention the same concept, $c$, call it $\tilde{\mathcal{R}}_{g}^{+}(I, Q, c)$, which we consider as candidate positive references for $Q$.
The set of questions not mentioning any of the same concepts are considered as candidate negative references, call it $\tilde{\mathcal{R}}_{g}^{-}(I, Q, c)$.
To increase the likelihood of the concept appearing in the image, we employ a set of NLP-based heuristics to remove questions whose images may not contain the concept, such as counting questions with an answer of ``$0$''.

\begin{figure}[t]
\begin{center}
\includegraphics[width=\linewidth]{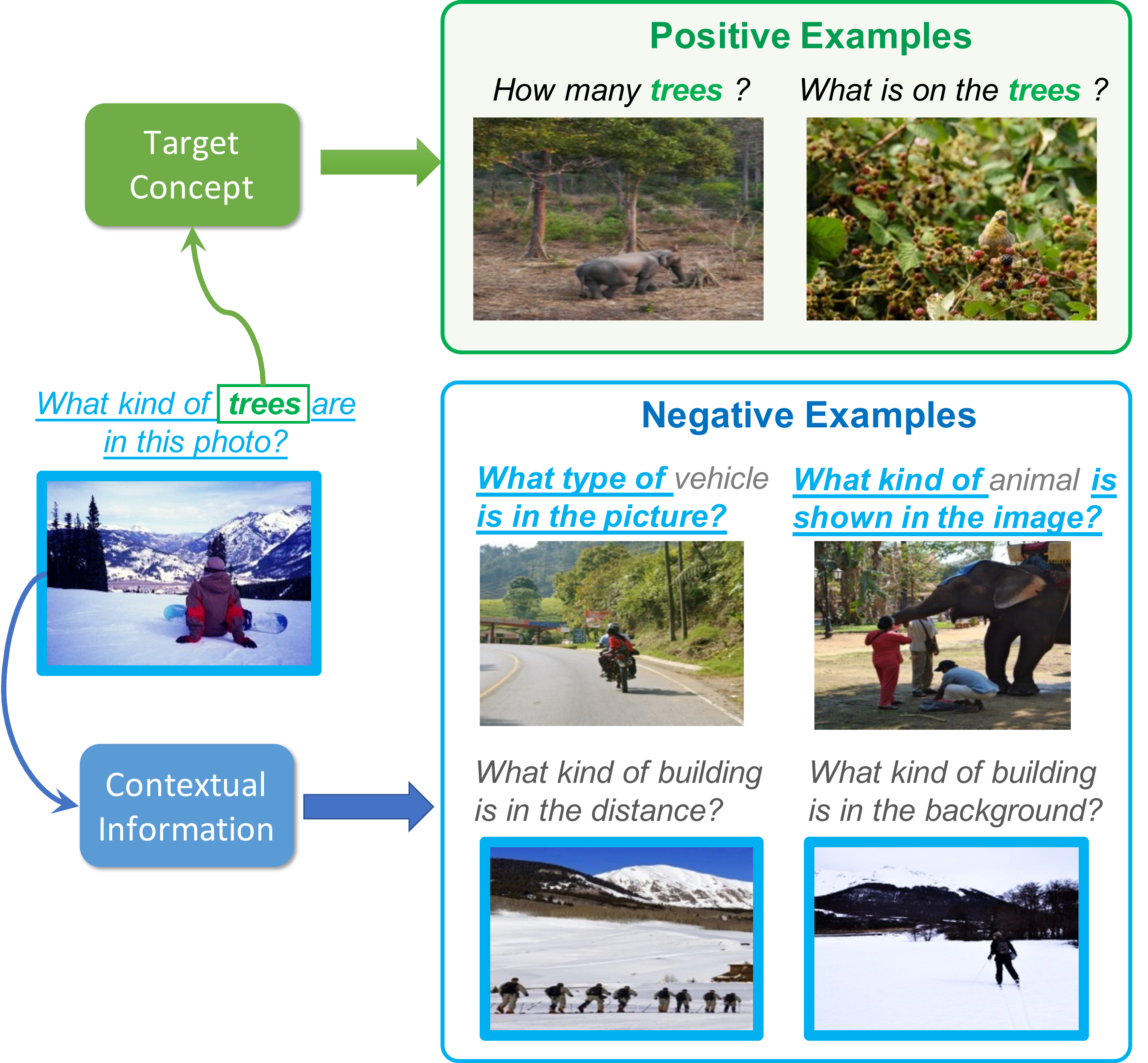}
\end{center}
\caption{Target example (left) and a reference set of positive/negative examples (right) from our CCC references.}
\label{fig:refset}
\end{figure}

\noindent\textbf{Concept-Context Contrastive (CCC) References.}
Given a target question, $Q$, and a target concept mention, $c$, in $Q$, we could simply create reference sets by randomly sampling positive and negative examples~\cite{akbari2019multi,oord2018representation} from $\tilde{\mathcal{R}}_{g}^{+}(I, Q, c)$ and $\tilde{\mathcal{R}}_{g}^{-}(I, Q, c)$, respectively, based solely on whether the question contains $c$ or not.
However, we propose a novel reference example filtering strategy to encourage concept grounding.
Our motivation is that, during VQA training, a concept often co-occurs with certain types of visual scenes or language priors.
So the positive and negative examples should force the model to not rely on superficial cues when contrasted against the target example and, instead, look at the correct visual regions.
Our solution is to build sets of refined reference candidates, $\mathcal{R}_{g}^{+}(I, Q, c)$ and $\mathcal{R}_{g}^{-}(I, Q, c)$, for each $(I, Q, c)$ tuple to ensure that the co-occurrence factor present in the dataset can be reduced.
As shown in \figref{fig:refset}, we want to find positive examples that also contain the concept ``\textit{tree}'', but with distinct visual scenes and questions from the target.
For negative examples, we seek distractors that are similar to the target in terms of the question or visual scene ({\sl e.g.,} mountains with skiers in \figref{fig:refset}), but do not reference ``\textit{tree}''.
To achieve this, we first represent the context of $c$ by masking out $c$ in the question and inputting the masked question and the image into off-the-shelf feature extractors to obtain question context representation $q$ and image representation $v$.\footnote{We use BERT~\cite{devlin2019bert} for questions and ResNet101~\cite{he2016resnet} for images.}
We measure the contextual similarity by:
\begin{align}\label{eq:refrank}
     \xi = \beta\text{cos}(q, q') + (1 -\beta)\text{cos}(v, v'),
\end{align}
where $\beta$ is a scalar and $(v, q)$ and $(v', q')$ are the representations from target and candidate examples, respectively.

To select positive examples from $\tilde{\mathcal{R}}_{g}^{+}(I, Q, c)$, we use $\beta=0.6$ and sample a set, $\mathcal{R}_{g}^{+}(I, Q, c)$, of $N^{+}$ examples that \emph{minimize} $\xi$ as our candidate positive examples for $(I, Q, c)$.
For negatives, we apply two settings of $\beta$ that \emph{maximize} $\xi$: $\beta=0.7$, which favors examples with more textual similarity, and $\beta=0.3$, which prioritizes images with similar visual context.
We select $N^{-}$ examples from each setting as our candidate negative examples, $\mathcal{R}_{g}^{-}(I, Q, c)$.
Illustrated in \figref{fig:refset}, when sampling reference sets from these two sets of candidates, the examples encourage the model to learn the specific correspondence between the concept mention in the question and its appearance in the image.
Intuitively, the model must to learn to ground the concept mention in the presence of the distractors.

\noindent\textbf{Concept Grounding Loss.}
Let $(I, Q)$ and $c$ be the target example and target concept mention, respectively, and let $\mathcal{X} = \{(I_k, Q_k)\}_{k=1}^{K}$ be a corresponding reference set.
Let $k^{*}$ be the index of the positive example in $\mathcal{X}$ sampled from $\mathcal{R}_{g}^{+}(I, Q, c)$, while the other $K - 1$ examples are negative examples from $\mathcal{R}_{g}^{-}(I, Q, c)$.
Let $w_i$ be the token in $Q$ that refers to the concept $c$.
We mask out $w_i$ and input this masked version of the question along with the corresponding image into the model, $f$, which outputs multi-modal representations from which we extract the representation of the masked concept token, $h_i$.
Next, we individually feed the examples from $\mathcal{X}$ into the model to obtain each token representation $\hat{h}_{k,j}$, where $j$ is the index of a token in $Q_k$.
Let $\hat{h}_{k^{*},j^{*}}$ be the representation of the concept mention in the positive example's question.
Our grounding loss is an NCE objective~\cite{chen2020nce,oord2018representation} that requires the model to match the multi-modal representation of the masked concept mention to the representation of the same concept mention in the reference set:
\begin{align}\label{eq:groundloss}
    \mathcal{L}_{g} = -\log \frac{\exp(\text{sim}(\phi_{g}(h_i), \phi_{g}(\hat{h}_{k^{*},j^{*}})))}{\sum_{k,j} \exp(\text{sim}(\phi_{g}(h_i), \phi_{g}(\hat{h}_{k,j})))} ,
\end{align}
where $\phi_{g}$ is a learned projection function and $\text{sim}(\cdot,\cdot)$ is a similarity function ({\sl e.g.,} dot product or cosine similarity).
To correctly match $h_i$ with $\hat{h}_{k^{*},j^{*}}$, the model must encode the visual features that match between the images of these examples in both token representations.
Our CCC references encourage these representations of the concept mention in the positive example and the masked concept mention to be grounded to the right visual regions as the model cannot rely on superficial textual or visual co-occurrences.

\subsection{Skill Matching}\label{sec:skillmatch}

Contrary to concepts, the essential skill needed to answer a certain question is largely independent of image appearances and mentions of concepts in the question.
For example, \emph{counting} questions should share a similar process to produce an answer: image areas associated with the subjects of counting are summarized to make the count prediction.
This process should be independent of the type of objects being asked about.
In other words, we seek to learn summary representations of questions that share the essential steps to infer the answer and are invariant to concepts.

\noindent\textbf{Skill References.}
A straightforward approach to learn skills is to annotate questions which explicitly require the same reasoning steps.
This annotation can be readily available on synthetic datasets~\cite{hudson2019gqa,johnson2017clevr}, but not available on datasets involving real-world images and questions.
Instead, we propose to mine sets of contrasting examples to learn which questions require the same/different skills, matching questions with the same skills.
Since the skills required for the question are typically indicated by the words of the question, we identify questions that are semantically similar.
Essentially, questions that require the same skill ({\sl e.g.,} ``\textit{What color ...}'') should be related to one another, regardless of the specific concept mentions in the question.
So, for each question, we mask out the concept words and we compute their BERT~\cite{devlin2019bert} representations.
For a given $(I,Q)$, the set of positive reference examples, $\mathcal{R}_{s}^{+}(I, Q)$, are sampled from the top-200 most similar questions using BERT representation, and the set of negative examples, $\mathcal{R}_{s}^{-}(I, Q)$, are randomly chosen from the rest of the dataset.

\noindent\textbf{Skill Matching Loss.}
For a given target example, $(I, Q)$, let $h$ be a summary representation of the target question.
This can be computed using a special input token like BERT~\cite{devlin2019bert} or via a pooling operation on all question token representations output from the encoder.
We sample a reference set of image-question pairs, $\{(I_l, Q_l)\}_{l=1}^{L}$, where the positive example, $Q_{l^{*}}$ from $\mathcal{R}_{s}^{+}(I, Q)$, shares the same skill as the target question, and the rest of the reference set are negative examples from $\mathcal{R}_{s}^{-}(I, Q)$.
Let $\hat{h}_l$ be a summary representation for a question in the reference set.
Shown in \figref{fig:architecture}, our skill matching loss is defined as
\begin{align}\label{eq:skillloss}
    \mathcal{L}_{s} = -\log \frac{\exp(\text{sim}(\phi_{s}(h), \phi_{s}(\hat{h}_{l^{*}})))}{\sum_{l} \exp(\text{sim}(\phi_{s}(h), \phi_{s}(\hat{h}_{l})))} ,
\end{align}
where $\hat{h}_{l^{*}}$ is the positive example representation and $\phi_{s}$ is another learned projection function.
This loss makes the representations of questions with the same skill more similar, regardless of the concepts mentioned.

\subsection{Training Procedure}\label{sec:train_procedure}

With our losses, we use a multi-tasking learning procedure~\cite{dong2015multi,luong2016multi}, where at each step we employ our objectives with probability $p_{\text{sep}}$ or not with probability $1 - p_{\text{sep}}$.
During training, we always first sample an instance from the labeled data, $\mathcal{D}^{a}$, and update the model by minimizing the VQA objective.
If at the current iteration we do not use our skill and concept objectives, then we only use the VQA objective.
Otherwise, we first use the VQA objective and then apply our other objectives.
Both objectives are computed in the same fashion:
for $\mathcal{L}_{g}$ (or $\mathcal{L}_{s}$), we sample a target example from $\mathcal{D}^{a}\cup\mathcal{D}^{u}$ along with $N_{r}^{+}$ positive examples from $\mathcal{R}_{g}^{+}$ (or $\mathcal{R}_{s}^{+}$) as well as $N_{r}^{-}$ negative examples from $\mathcal{R}_{g}^{-}$ (or $\mathcal{R}_{s}^{-}$), combine the sampled references to form the current reference set, and compute the loss term.
We then sum $\mathcal{L}_g$ and $\mathcal{L}_{s}$, and update the model to minimize the negative sum.

\begin{table*}[!ht]
\footnotesize
  \begin{center}
    \begin{tabular}{c|c|c|c|c|c|c|c|c}
    \hline
    
    \multirow{2}{*}{\textbf{Model}} & \multicolumn{4}{c}{Counting} & \multicolumn{2}{|c}{Color} & \multicolumn{1}{|c|}{Subcat.} & \multirow{2}{*}{\textbf{Overall}} \\ 
    \cline{2-8}

     & animal & \{animals\} & \{vehicles\} & \{electronics\}
     & animal  & \{dishware\}
     & vegetable &  \\
    
    \hline

     XNM~\cite{shi2019xnm} & 56.02 & 48.32& 44.35 & \underline{51.94}
     & 77.22
     & \underline{65.73} & 57.33 & \underline{57.27} \\
     
     StackNMN~\cite{hu2018explainable}  & \underline{54.22} & \underline{47.56} & 46.10 & 52.83
     & 76.57
     & 69.22 & \underline{57.17} & 57.67 \\
     
     X-Att~\cite{tan2019lxmert}  & 58.94 & 56.28 & 46.30 & 57.05
     & \underline{73.15}
     & 67.29 & 57.25 & 59.47\\

     X-BERT~\cite{chen2020uniter}  & 63.58 & 54.58 & \underline{42.34} & 56.84
     & 75.88
     & 70.31 & 58.96 & 60.36 \\
     
     \hline
     
     Base  & 62.57 & 59.19 & 48.33 & 61.84
     & 76.57 & 72.91
     & 58.33 & 62.82\\
     
     Ours & \textbf{65.16} & \textbf{59.87} & \textbf{50.75}  & \textbf{62.21}
     & \textbf{77.45} & \textbf{73.76}
     & \textbf{61.04} & \textbf{64.32}\\

    \hline
    \end{tabular}
  \end{center}
\caption{VQA accuracy on novel skill-concept compositions. The highest and the lowest numbers of each experiment are emphasized.}
\label{tab:newconceptskill}
    
\end{table*}

\begin{table*}[ht]
\footnotesize
  \begin{center}
    \begin{tabular}{c|c|c|c|c|c|c|c|c|c|c|c}
    \hline
    \textbf{Model} & lamp
     & fruit & fridge & surfer 
     & flag & skateb. & oven 
     & sheep & banana & zebra & \textbf{Overall} \\
    \hline 
     XNM~\cite{shi2019xnm} & 53.69
     & 50.23 & 57.98 & 72.68
     & 36.58 & 70.16 & 53.49
     & 54.96 & \textbf{52.35} & 61.50 & 56.36 \\
     
     StackNMN~\cite{hu2018explainable} & 54.27
     & 46.10 & 58.97 & 74.10
     & \textbf{41.31} & 74.11 & 56.30
     & 57.12 & 50.98 & 61.25 
     & 57.45 \\

     X-Att~\cite{tan2019lxmert} & 46.26
     & 33.10 & 51.52 & 67.68
     & \underline{31.53} & 69.73 & 51.69
     & \underline{49.83} & \underline{41.71} & \textbf{64.93} 
     & 50.80\\
     
     X-BERT~\cite{chen2020uniter} & \underline{44.43} & \underline{30.72} & \underline{50.83} & \underline{61.60} & 32.46 & \underline{66.05} & \underline{48.10} & 50.32 & 43.10 & 57.06 & \underline{48.47} \\
     
     \hline
     
     Base & 55.14
     & 52.99 & 59.06 & 74.12
     & 39.05 & 71.67 & 56.60
     & 63.31 & 49.83 & \underline{56.05 }
     & 57.78\\

    Ours & \textbf{57.40}
     &\textbf{54.40}&\textbf{60.92}& \textbf{74.36}
     &40.15&\textbf{75.27}& \textbf{59.91}
     &\textbf{64.04}&50.78&60.77 
     & \textbf{59.80}\\
     \hline
     
    \end{tabular}
\end{center}
\caption{VQA Accuracy on individual novel-concept split. \emph{skateb.} refers to \emph{skateboarder}. }
\label{tab:newconcept}
\end{table*}

\section{Experiments}
\label{sec:exp}
\noindent\textbf{Data and Settings.}
We run our experiments on VQA v2~\cite{goyal2017vqav2}, which contains real images, human-written questions, and a variety of skills required to answer the questions.
Since the goal of this work is to examine a model's performance on different types of novel questions, it requires the availability of answer annotations for the test data.
Since the annotations of test-dev and test-std sets of VQA v2 are not publicly available, we use questions from the validation set for testing as is common~\cite{agrawal2018vqacp,cycleconsist2019}.
We do not train or tune hyperparameters with the validation set; it is strictly used for evaluation.
We compare performance using the VQA accuracy~\cite{antol2015vqa} on different splits of novel questions.
Details are provided in \appref{appendix:implementation}.

\noindent\textbf{Model Comparisons.}
We select a set of recent VQA models to benchmark their novel-VQA performance.
The first category is compositional models~\cite{andreas2016neural,hu2018explainable,hu2017n2nnmn,shi2019xnm}.
We use \textbf{StackNMN}~\cite{hu2018explainable} and \textbf{XNM}~\cite{shi2019xnm}, which are designed to handle compositional questions, like those in CLEVR~\cite{johnson2017clevr}, and have state-of-the-art performance on these datasets while also being applicable to real images without supervision from functional programs or image scene graphs.

The second type of model we experiment with is transformer-based~\cite{chen2020uniter,devlin2019bert,li2020unicoder,li2020oscar,lu2019vilbert,tan2019lxmert,vaswani2017attention,yu2019mcan}.
We use two top-performing transformer architectures from this model family: 1) a two-stream, cross-attention model~\cite{tan2019lxmert} (\textbf{X-Att}), which has modality specific branches and cross-attentions in early layers followed by multi-modal layers later in the network; and 2) a vision-and-language transformer model~\cite{chen2020uniter} (\textbf{X-BERT}) that acts as multi-modal encoder throughout the entire network.
For fair comparison, we do not use pre-training, same as our model, since we are specifically interested in the generalization ability of data-efficient models without requiring large-scale ({\sl e.g.,} 9M+ image-text pairs), in-domain data external to VQA~\cite{hendricks2021decoupling,singh2020pretrainright}.

Lastly, for our base model, we employ a variant of the standard multi-modal transformer where image features can be attended by both the CLS token and language features.
When the base model is trained without the proposed skill-concept contrastive losses, it serves as a baseline model (denoted by \textbf{Base}).
Details are in \appref{appendix:implementation} and \appref{appendix:architecture}.

\subsection{Novel Skill-Concept Composition VQA}\label{sec:resultcomp}

We select three prevalent and common skills present in \ogvqa: counting, color querying and subcategory recognition.
For each skill, we remove the data labels for its co-occurring questions with one concept or a set of multiple concepts which can form a distinct category from training, and then test on these compositions.
The concepts (or concept groups) are randomly sampled with two criteria: each skill-concept composition contains reliable amount of test data to measure accuracy and the compositions have diverse coverage across the dataset (more details in \appref{appendix:skillconcept}).

\tabref{tab:newconceptskill} shows the VQA accuracy on each of the novel 
compositional subset. Interestingly, although neural module networks are designed
to explicitly break down the question answering process into subtasks, which
in principle should help with adapting these subtasks to new questions and thus
generalize better, they yield lower performance than transformer models.
This may be due to the effective feature learning capacity of self-attention mechanisms.
Among all transformer models, our base encoder achieves competitive performance to existing networks, demonstrating that it is a strong baseline among multi-modal transformers.
Finally, our contrastive learning framework outperforms the baseline and all other approaches across each novel composition set.
This supports the effectiveness of our framework for generalization to new compositions.

\subsection{Novel-Concept VQA}

For this experiment, we are interested in the setting where models are never trained to answer questions about a concept but can make use of the unlabeled image-question pairs, and then are tested on questions that have mentions of this given concept.
Similar to the previous experiment, these concepts were sampled to maximize coverage as well as maintaining a reasonable test size.
This setting is more challenging than the previous experiment since the model misses the VQA training supervision on any questions that have the given concept, as opposed to any questions that have both the given concept and a certain skill.

We provide qualitative examples of novel-concept VQA in \appref{appendix:examples} and
report quantitative results in \tabref{tab:newconcept}. 
For this more challenging setting of novel question answering, on average, two of the existing transformer architectures underperform other models by a noticeable margin.
This may suggest that the transformer architectures, which perform well on large-scale vision and language pre-training, may have difficulty specializing to the VQA task.
The Base model slightly outperforms neural module networks.
Lastly, our framework again outperforms all models on average, demonstrating its value in improving VQA generalization ability on novel concepts.

\subsection{Analysis}

\begin{figure}[ht]
\begin{center}
\includegraphics[width=\linewidth]{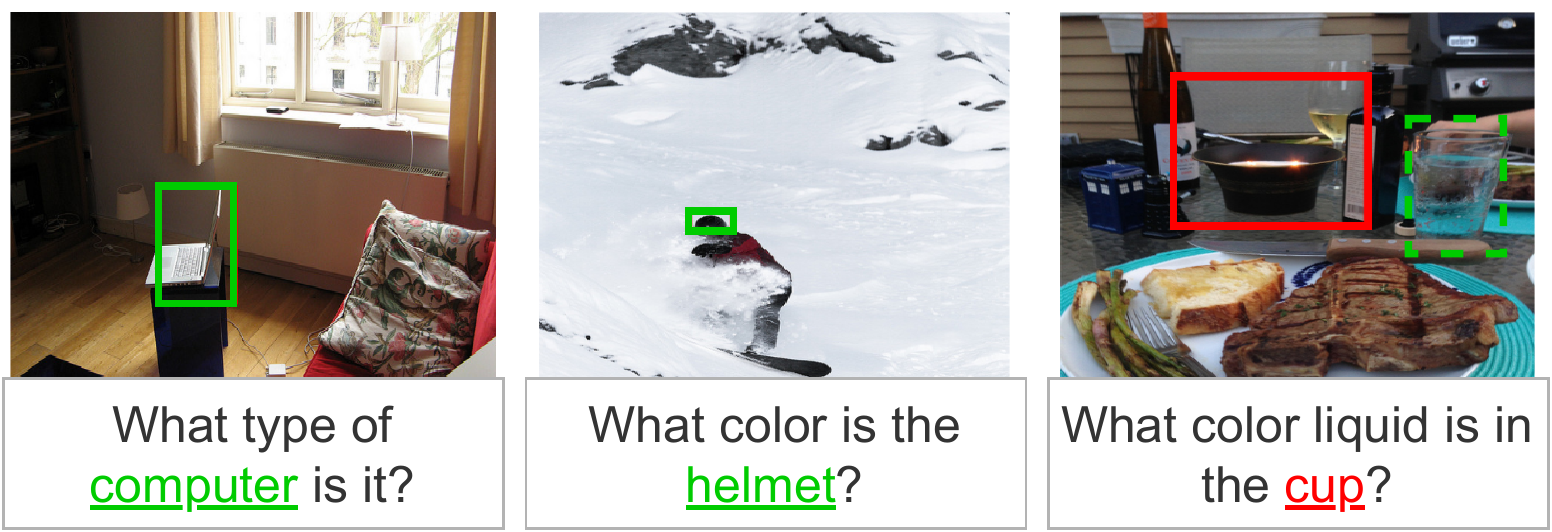}
\end{center}
\caption{\textcolor{ForestGreen}{Correct} and \textcolor{red}{incorrect} grounding examples. We visualize the most similar visual region to the concept in the question.}
\label{fig:ground_output}
\end{figure}

\noindent\textbf{Concept Grounding.}
Since our approach learns to ground concepts without strong supervision, we would like to test its grounding abilities directly.
To obtain an evaluation set, we manually annotate 320 image-question-concept tuples with the visual regions in the image that corresponds to the concept in each tuple.
Candidate visual regions are found using Faster-RCNN~\cite{ren2015faster}.
We use recall@5 as our grounding metric, considering a grounding correct if the correct visual region falls within the top 5 most similar visual regions to the target concept token.
The model trained with our framework achieves a grounding recall of \textbf{59.12}, compared to \textbf{43.71} of Base.
Note that our framework obtained this improvement with no additional training data for grounding.
As shown in \figref{fig:ground_output}, our model can often correctly ground a variety of objects, but can be fooled by ambiguous looking concepts like the candle in the incorrect example.
Further, it is challenging to learn to differentiate concepts that almost always co-occur ({\sl e.g.,} ``\textit{shirt}'' and ``\textit{person}'').

\noindent\textbf{Loss Ablation.}
We ablate our losses by sampling three novel compositions and three novel concepts and report their average performance in \tabref{tab:ablateLoss}.
Adding our losses leads to consistent gains, with top performance achieved with our full framework.
When used alone, our grounding loss seems to contribute a larger benefit compared to the skill loss.
Nonetheless, the best performance is achieved by combining the two components, further supporting the value of skill and concept separation.
We also experiment with a masked language modeling (MLM) objective~\cite{devlin2019bert} that replaces our losses.
Our objectives perform better than the MLM objective, implying that the improvements our objectives offer are not simply due to additional data.

\noindent\textbf{CCC Reference Sets.}
To study the effects of our CCC reference set selection strategy, we compare it with the commonly used random sampling method~\cite{akbari2019multi,oord2018representation} and report novel-concept VQA results in \tabref{tab:randomrefset}.
We train both models with our full framework, the only difference being the reference set construction method for the concept loss. 
Both models improve upon the Base model, with our reference set construction method offering more consistent gains.

\noindent\textbf{Existing Benchmarks.}
We also evaluate on VQA-CP~\cite{agrawal2018vqacp} and the test-dev/test-std splits of \ogvqa (\tabref{tab:skillconept_stdvqa}).
While we see gains in general, notably, our approach is able to improve on VQA-CP without extra annotations, ensembling/tuning, or a performance drop on \ogvqa.

\begin{table}[t]
\footnotesize
\begin{center}
\begin{tabular}{c|c|c} 
 \hline
 \textbf{Model} & Avg. Novel Count & Avg. Novel Concepts \\ 
 \hline 
 Base & 58.03 & 60.42 \\
 Base+MLM & 58.41 & 60.25 \\
 Base+$\mathcal{L}_{s}$ & 58.83 & 61.85 \\
 Base+$\mathcal{L}_{g}$ & 59.80 & 62.06 \\
 Ours & \textbf{60.71} & \textbf{63.19} \\ 
 \hline
\end{tabular}
\end{center}
\caption{Effect of using different losses on novel skill-concept composition and novel-concept VQA.}
\label{tab:ablateLoss}
\end{table}

\begin{table}[t]
\footnotesize
\begin{center}
\begin{tabular}{c|c|c|c|c|c} 
 \hline
 \textbf{Model} & lamp & fruit & fridge & surfer & flag \\ 
 \hline 
Base & 55.14 & 52.99 & 59.06 & 74.12&  39.05\\
\hline
Random & \textbf{+2.80} &+0.49  & -0.16 & -0.14& +0.50\\
CCC (Ours) & +2.26 & \textbf{+1.41}& \textbf{+0.98} & \textbf{+0.24}&  \textbf{+1.10}\\
 \hline
\end{tabular}
\end{center}
\caption{Comparing different reference set construction schemes for concept learning across five different concepts.}
\label{tab:randomrefset}
\end{table}

\begin{table}[t]
\footnotesize
\begin{center}
\begin{tabular}{c|c|c|c}
 \hline
 \textbf{Model} & VQA-CP & Test-dev & Test-std \\ 
 
 \hline 
 Base & 40.98 & 69.60 & 69.99 \\
 Ours & \textbf{41.71} & \textbf{69.78} & \textbf{70.09} \\ 
 \hline
\end{tabular}
\end{center}
\caption{Single model VQA performance on VQA-CP~\cite{agrawal2018vqacp} and \ogvqa test-dev and test-std splits. Both models see the exact same training data (no compositions/concepts are removed).}
\label{tab:skillconept_stdvqa}
\end{table}

\section{Conclusions}

We propose a new setting for generalization in VQA: measuring the ability to compose the skills needed to answer a question and the visual concepts that should be grounded to the image.
We show that existing approaches have difficulty generalizing to unseen compositions of these two factors.
We present a novel approach that implicitly disentangles skills and concepts, while grounding concepts visually, using a contrastive learning procedure.
Our approach is able to learn from unlabeled VQA data in order to answer questions about previously unseen concepts.
Results on the \ogvqa show that the proposed framework can achieve state-of-the-art performance on novel skill-concept compositions as well as generalize from unlabeled data.

\smallskip
{

\noindent\textbf{Acknowledgements:}
We thank David Cox for the helpful discussions.
From the UIUC side: This work was in part supported by the U.S. DARPA AIDA Program No. FA8750-18-2-0014. The views and conclusions contained in this document are those of the authors and should not be interpreted as representing the official policies, either expressed or implied, of the U.S. Government. The U.S. Government is authorized to reproduce and distribute reprints for Government purposes notwithstanding any copyright notation here on.

}

\clearpage

{\small
\bibliographystyle{ieee_fullname}
\bibliography{refs}
}

\newpage
\clearpage
\begin{center}
{\bf {\Large Appendix\\} }
\end{center}
\appendix

\section{Skill and Concept Details}\label{appendix:skillconcept}

To construct a comprehensive list of common skills required to answer a VQA question,
we draw information from three sources: (1) our own annotation on 400 randomly selected VQA questions; (2) user study from~\cite{skillsvqa2020}; and (3) previous work on question types~\cite{hudson2019gqa,kafle2017analysis}. The user study in 
~\cite{skillsvqa2020} only provides four types of vision skills. Existing work on
question types have relevant information, however, the question types are not always
directly translatable to our paradigm of skill and concept composition. For example,
concept recognition is considered as a question type in \cite{kafle2017analysis} (object presence), but in our
framework, it is considered as \emph{concept grounding} rather than as a separate skill.
Besides, existing question types are sometimes
incomplete~\cite{kafle2017analysis}, or not representative
of natural questions typically asked about images~\cite{hudson2019gqa}. For instance, skills that require comparison or text reading
form $\sim6\%$ of the questions according to our labeling results, but they are not
covered in~\cite{kafle2017analysis}. We consolidate our annotations with
groupings in existing work, which results in the following set of skills:

\begin{itemize}
\small
\setlength\itemsep{-0.1em}
\item Color recognition: \textit{What color hair does the woman have? What color is his shirt?}

\item Attribute recognition (non-color attributes):\ \textit{Is the bed made? Is this desk messy?}

\item Subcategory recognition: \textit{What kind of car is parked? What kind of animals are shown?}

\item Action recognition: \textit{What is the man doing in the street? Are they comparing their phones?}

\item Scene recognition: \textit{Is this on a farm? Are they outside?}

\item Counting: \textit{How many lights are there? How many zebras are in this picture?}

\item Commonsense knowledge: \textit{Is the sun going down? Is this in America?}

\item Positional reasoning: \textit{What is on top of the toaster? What is the zebra standing on? }

\item Text Recognition: \textit{What number bus is it? What is the store called?}

\item Comparison: \textit{Is the tank the same color as the toilet? Are they facing the same direction?}

\end{itemize}

\begin{figure}[!ht]
\begin{center}
\includegraphics[width=\linewidth]{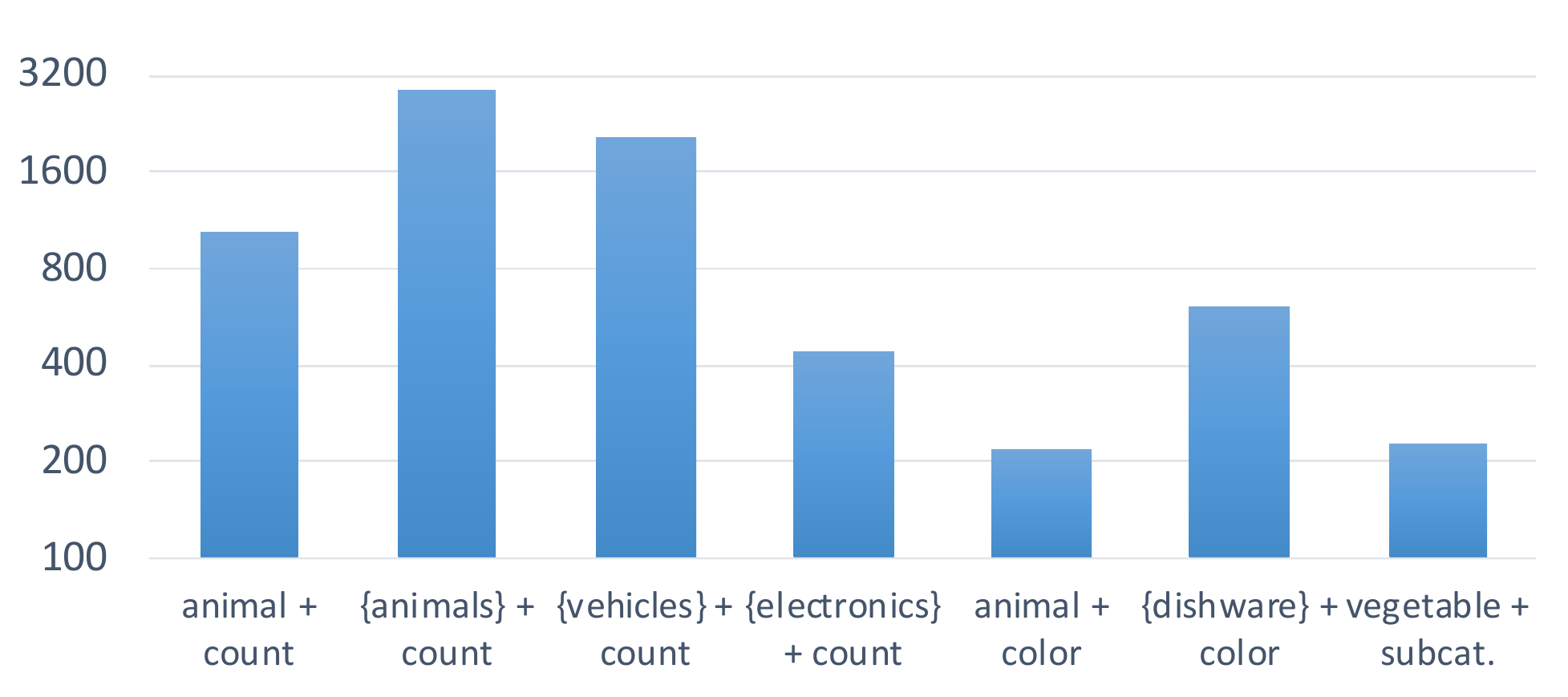}\\
\includegraphics[width=\linewidth]{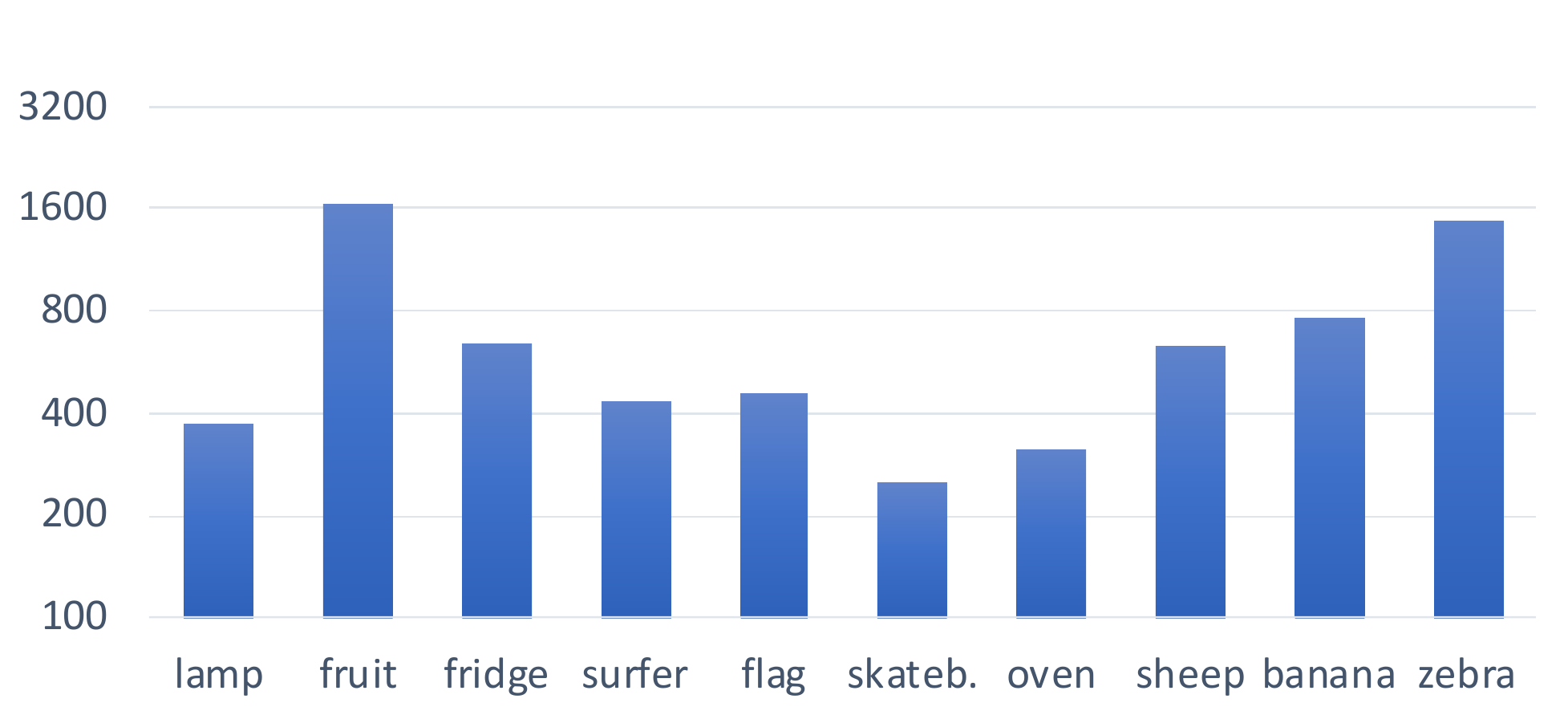}
\end{center}
\caption{Novel skill-concept composition (top) and novel concept (bottom) question statistics.}
\label{fig:stats}
\end{figure}

We also provide additional information and statistics of the novel compositions.
To facilitate further research on novel-VQA evaluation, we will provide concept and skill annotations, and the respective data indices for each set of novel composition. 
The list of concepts within each concept group is:
\begin{itemize}
\small{
\setlength\itemsep{0em}
\item \{animals\}: \textit{giraffe}, \textit{zebra}, \textit{bird}, \textit{sheep}, \textit{horse}, \textit{elephant}, \textit{cow}, \textit{dog}, \textit{cat}

\item \{vehicles\}: \textit{motorcycle}, \textit{airplane}, \textit{plane}, \textit{jet}, \textit{bus}, \textit{car}, \textit{truck}, \textit{bike}, \textit{bicycle}

\item \{electronics\}: \textit{computer}, \textit{monitor}, \textit{laptop}, \textit{phone}, \textit{cellphone}

\item \{dishware\}: \textit{plate}, \textit{bowl}
}

\end{itemize}

The list of sizes for each novel testing split is shown in \figref{fig:stats}.
To determine the compositions/concepts that we use, we employ a few criteria:
1) each skill-concept composition (or concept) must have a minimum of 400 training questions and 200 testing questions;
2) for compositions, to increase coverage and ensure the minimally
required size, we use concept groups where the concepts in a group all fall under a broader category ({\sl e.g.,} \{animals\} = \{\textit{giraffe}, \textit{zebra},...\}).
We then sample from these compositions/concepts to conduct experiments on.

\section{Approach Details}\label{appendix:approach_details}

Here, we detail the projection functions, similarity functions, and other settings for our approach.
In the following equations, all $W$ and $b$ are learned parameters.

\noindent\textbf{Concept Grounding.}
For our concept grounding loss, we want to maximize the similarity of the masked target concept token to the correct concept token in the positive reference example.
Since we are directly comparing tokens between examples, we model the similarity computation as an attention~\cite{bahdanau2015neural,luong2015attention,vaswani2017attention}  with which the model must point~\cite{vinyals2015pointer} to the correct concept token.
Specifically, our projection function, $\phi_g(\cdot)$, and similarity function, $\text{sim}(\cdot,\cdot)$, are defined as
\begin{align}\label{eq:groundproj}
    \phi_g(x) &= W_{g} x + b_g \\
    \text{sim}(x, y) &= \frac{x^\intercal y}{\sqrt{d}} ,
\end{align}
where $d$ is the dimension of $x$ and $y$, $W_g\in\mathbb{R}^{d\times d}$, and $b_g\in\mathbb{R}^{d}$.
Though this is similar to an attention, our formulation matches more traditional contrastive learning objectives~\cite{chen2020nce,oord2018representation}, where $\sqrt{d}$ is the temperature and we use a dot product as our similarity measure.

\noindent\textbf{Skill Matching.}
Our skill matching loss seeks to maximize the similarity of the summary representation of the target question with the summary representations of other questions with the same skill.
To obtain summary representations of questions, we simply use mean pooling over the question token representations.
We define our projection function, $\phi_{s}(\cdot,\cdot)$, and similarity function, $\text{sim}(\cdot,\cdot)$, as
\begin{align}\label{eq:skillproj}
    \phi_s(x) &= W_{s}^{(2)}\psi(W_{s}^{(1)} x + b_{s}^{(1)}) + b_{s}^{(2)} \\
    \text{sim}(x, y) &= \frac{\cos(x, y)}{\tau_s} ,
\end{align}
where $\psi$ is a ReLU nonlinearity and $\tau_s$ is a temperature ($\tau_s = 0.5$ in our experiments).
Since we are not directly comparing token representations, we use the more standard contrastive objective~\cite{chen2020nce} as opposed to the attention-based formulation used for concept grounding.

\noindent\textbf{Reference Sets and Training Procedure.}
When forming our CCC candidate references from which we sample our reference sets, we use $N^{+}=20$ and $N^{-}=40$ (since we have two settings for negative examples), so there are $N^{+}$ positive and $N^{-}$ negative examples  that can be selected from to form a reference set for a given target example.
Meanwhile, we use $N^{+}=200$ and $N^{-}=200$ for our skill matching candidate references.
Then, in our multi-task training procedure, we use $p_{\text{sep}} = 0.1$ as the probability of applying our framework at each training step.
Additionally, we simply use $N_{r}^{+} = 1$ and $N_{r}^{-} = 2$ for both concept grounding and skill matching, so the model will contrast between a single positive example and two distractor negative examples.
For our concept grounding loss, we sample one negative example from both of our settings as our negative examples.

\section{Experimental Details}\label{appendix:implementation}

\subsection{Dataset Information}

We use \ogvqa~\cite{goyal2017vqav2} for our main experiments.
For training, we only use the training split.
Since the testing data of \ogvqa~\cite{goyal2017vqav2} does not have public groundtruth information, we use the validation split of \ogvqa as the testing set for novel-VQA.
To form our novel skill-concept and novel-concept VQA test splits, we automatically label questions with the skills and concepts using different NLP-based rules.
For labeling skills, we use question template matching ({\sl e.g.,} ``\textit{How many ...}'') as well as verifying that the answers fit the matched templates.
For labeling concepts, we utilize lemmatization and POS taggging and collect the frequent nouns.
We then create different training splits that have a specific skill-concept composition or concept removed.

We also run experiments on the test-dev, test-std, and VQA-CP~\cite{agrawal2018vqacp} splits of \ogvqa.
When evaluating on test-dev and test-std, we train on the validation set and additional Visual Genome data~\cite{teney2018tips}.

\subsection{Model Configurations}

All models use the same visual features~\cite{anderson2018butd}.\footnote{\url{https://github.com/peteanderson80/bottom-up-attention}}
We also use GloVe word embeddings~\cite{pennington2014glove}.\footnote{Common Crawl 840B: \url{https://nlp.stanford.edu/projects/glove/}}
Our baselines from prior work follow the recommended settings provided by the authors, whenever possible.

For XNM~\cite{shi2019xnm}, we use the implementation provided by the authors as well as the recommended settings.\footnote{\url{https://github.com/shijx12/XNM-Net}}
To ensure consistency between the two compositional models, we implement StackNMN~\cite{hu2018explainable} within the same code base as XNM.
Specifically, we match the controller and the modules of StackNMN to the original paper.
We use hidden dimension sizes of 512 for StackNMN and 1024 for XNM.
We use the recommended number of reasoning steps, $T = 3$, for XNM and use the same for StackNMN.
Both these models are trained with the Adam optimizer~\cite{kingma2014adam} and have the same learning rate of 0.0008 and batch size of 256.

For both X-Att~\cite{tan2019lxmert}\footnote{\url{https://github.com/airsplay/lxmert}} and X-BERT~\cite{chen2020uniter}\footnote{\url{https://github.com/ChenRocks/UNITER}}, we use the original model source code.
For fair comparison, we do not use large-scale pre-training, same as our model.
For X-Att, we use the recommended settings with a hidden size of 768, 12 layers, and 12 attention heads.
X-Att uses the recommended learning rate of 0.0001, batch size of 64, 20 training epochs, and the Adam optimizer~\cite{kingma2014adam}.
Due to their similarities in architecture, we use the same settings for Base, X-Bert and our framework for a more head-to-head comparison.
Specifically, we use a hidden size of 512, 6 layers, and 8 attention heads.
We match the training settings as well: a learning rate of 0.0001, batch size of 64, 13 training epochs, step learning rate decay with a rate of 0.2, and the Adam optimizer~\cite{kingma2014adam}.

\section{Base Model Architecture}\label{appendix:architecture}

The base model (Base), to which we apply our framework, is based on the standard transformer encoder~\cite{chen2020uniter,devlin2019bert} with a few modifications.
As is standard with transformers, we input visual regions, question tokens, and a special \textsc{CLS} that is appended to the beginning of the inputs, which we use to predict answers via a softmax output layer.
There are two minor differences between a standard transformer and our base model:
First, before inputting the question into the tranformer layers, we encode sequential information in the question tokens using an bi-directional LSTM, yielding a slight improvement than positional embeddings~\cite{vaswani2017attention}.
Second, in each layer, the \textsc{CLS} token and visual regions can attend to all inputs, including themselves, and the question tokens only directly attend to themselves and the \textsc{CLS} token.
The change allows the \textsc{CLS} token to act as a bottleneck through which textual information interacts with the visual information.

\section{Qualitative Examples}\label{appendix:examples}

\begin{figure*}[th!]
\begin{center}
\includegraphics[width=\textwidth]{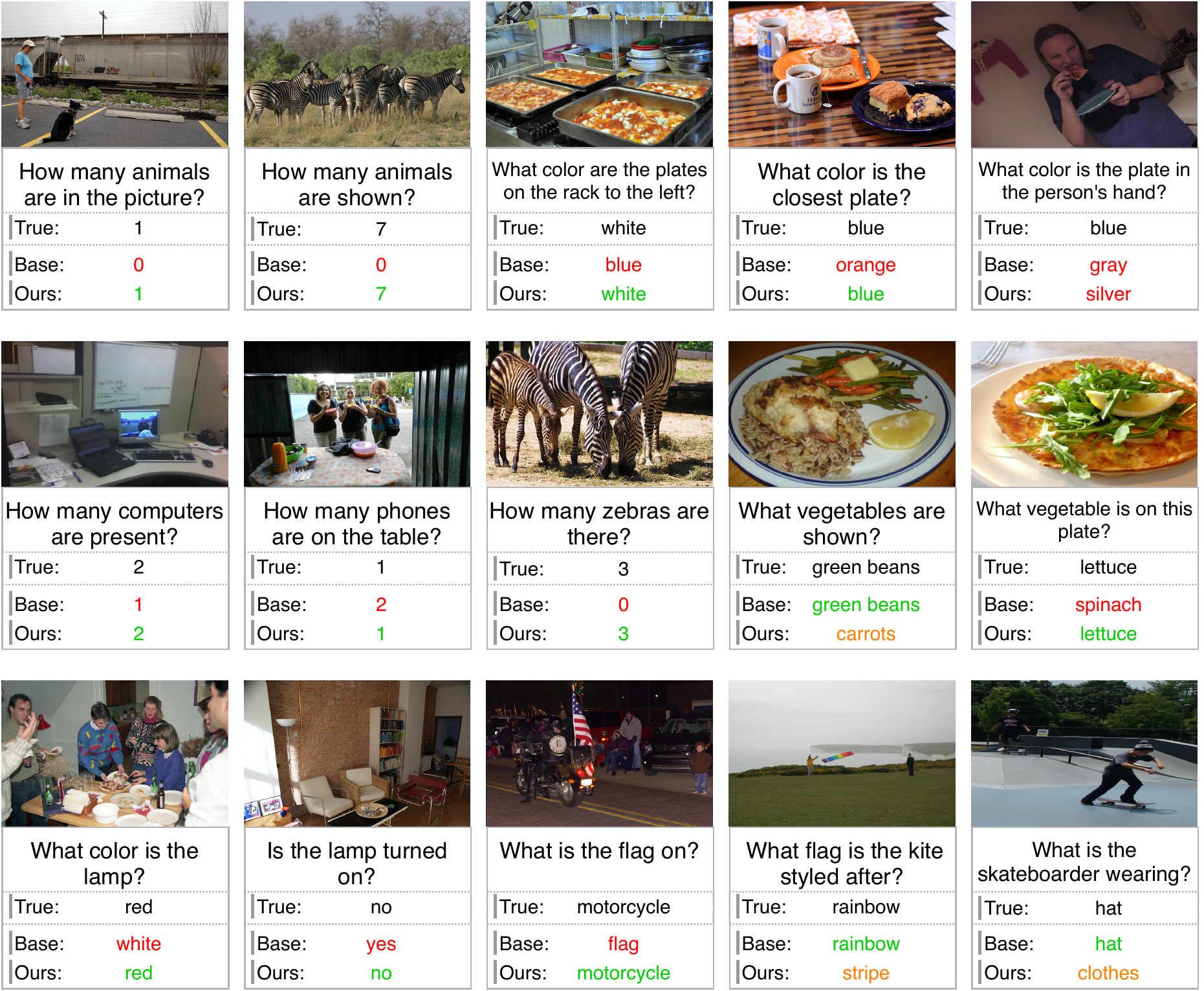}
\end{center}
\caption{\textcolor{ForestGreen}{Correct}, \textcolor{red}{incorrect}, and \textcolor{orange}{plausible} VQA output examples for novel skill-concept composition VQA (rows 1 and 2) and novel concept VQA (row 3), comparing the predictions of our approach (Ours) and the Base model.}
\label{fig:output_examples}
\end{figure*}

We show VQA output examples in \figref{fig:output_examples} that compare the performance of our approach versus Base, where the first two rows show predictions on novel skill-concept compositions and the last row shows predictions on novel concept VQA.
As a reminder, the models tested here never see labeled image-question pairs with the specific compositions/concepts during training.
Our approach allows the model to adapt to these unseen compositions.
We see that, for unseen compositions of \emph{counting} and different concepts, the base model struggles to recognize and count these concepts.
For example, we observe that despite the clear appearance of the animals in the images, the Base model is unable to transfer the skill of counting, whereas the model trained with our framework is able to handle these cases.
Similarly, in the third and fourth examples of the first row, we see an interesting effect where our approach is able to more precisely locate the specific ``\textit{plate}'' being referred to.
Another interesting example of the improvements that our grounding framework can offer is shown in the first three examples of the last row, where our model is able to locate the specific object and produce the correct answer.
The last two examples of the third row show some intriguing failure cases, where our model produces plausible yet somewhat generic answers compared to the baseline.

\end{document}